\documentclass[conference]{IEEEtran}
\IEEEoverridecommandlockouts
\usepackage{cite}
\usepackage{amsmath,amssymb,amsfonts}
\usepackage{algorithmic}
\usepackage{algorithm}

\makeatletter
\newcommand{\removelatexerror}{\let\@latex@error\@gobble}
\makeatother
\usepackage{graphicx}
\usepackage{textcomp}
\usepackage{xcolor}
\usepackage{caption}
\usepackage{enumerate}
\usepackage{booktabs} 
\usepackage{multirow}
\usepackage{textcomp,mathcomp}
\usepackage{subfig}
\usepackage{graphicx}
\usepackage{epstopdf}
\usepackage{epsfig}
\usepackage[colorlinks=true,linkcolor=blue]{hyperref}

\usepackage{hyperref}
\hypersetup{
    colorlinks=true,
    linkcolor=black,
    filecolor=black,      
    urlcolor=black,
    citecolor=cyan,
}

\def\BibTeX{{\rm B\kern-.05em{\sc i\kern-.025em b}\kern-.08em
    T\kern-.1667em\lower.7ex\hbox{E}\kern-.125emX}}
\begin{document}

\title{CyFormer: Accurate State-of-Health Prediction of Lithium-Ion Batteries via Cyclic Attention\\
}

\author{\IEEEauthorblockN{Zhiqiang Nie\thanks{*Equal contribution.}$^{1*}$, Jiankun Zhao$^{2*}$, Qicheng Li$^{2\dag}$, Yong Qin\thanks{$^{\dag}$Corresponding authors.}$^{2\dag}$}
\IEEEauthorblockA{$^{1}$ College of Cyber Science, Nankai University, Tianjin, China}
\IEEEauthorblockA{$^{2}$ College of Computer Science, Nankai University, Tianjin, China}
\IEEEauthorblockA{\{\href{mailto: 2012307@mail.nankai.edu.cn}{2012307}, \href{mailto: 2010535@mail.nankai.edu.cn}{2010535}\}@mail.nankai.edu.cn, 
\{\href{mailto: liqicheng@nankai.edu.cn}{liqicheng}, \href{mailto: qinyong@nankai.edu.cn}{qinyong}\}@nankai.edu.cn}
}

\maketitle
\hypersetup{hidelinks} 

\begin{abstract}
Predicting the State-of-Health (SoH) of lithium-ion batteries is a fundamental task of battery management systems on electric vehicles. It aims at estimating future SoH based on historical aging data. Most existing deep learning methods rely on filter-based feature extractors (e.g., CNN or Kalman filters) and recurrent time sequence models. Though efficient, they generally ignore cyclic features and the domain gap between training and testing batteries. To address this problem, we present CyFormer, a transformer-based cyclic time sequence model for SoH prediction. Instead of the conventional CNN-RNN structure, we adopt an encoder-decoder architecture. In the encoder, row-wise and column-wise attention blocks effectively capture intra-cycle and inter-cycle connections and extract cyclic features. In the decoder, the SoH queries cross-attend to these features to form the final predictions. We further utilize a transfer learning strategy to narrow the domain gap between the training and testing set. To be specific, we use fine-tuning to shift the model to a target working condition. Finally, we made our model more efficient by pruning. The experiment shows that our method attains an MAE of 0.75\% with only 10\% data for fine-tuning on a testing battery, surpassing prior methods by a large margin. Effective and robust, our method provides a potential solution for all cyclic time sequence prediction tasks. 
\end{abstract}


\section{Introduction}
\begin{figure}[htbp]
\centerline{\includegraphics[width = 0.5\textwidth]{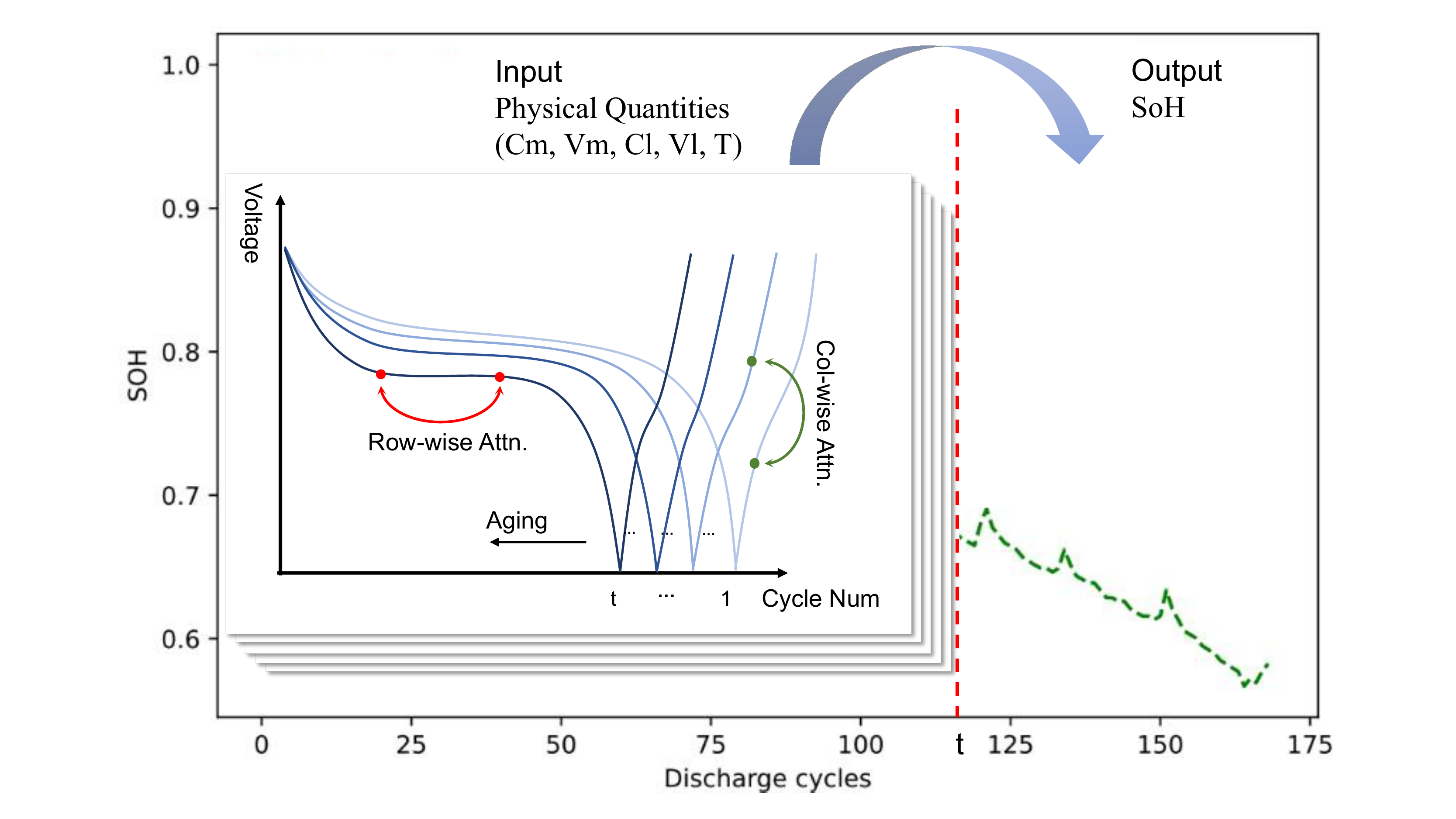}}
\caption{Illustration of the SoH prediction task and the row-wise and column-wise attention mechanism. The model takes in a series of historical physical quantities, and outputs SoH values of future cycles. In the experiment, the physical quantities we use are measured current (Cm), measured voltage (Vm), load current (Cl), load voltage (Vl) and temperature (T). Row-wise attention captures intra-cycle connections, while col-wise attention captures inter-cycle connections.}
\label{fig:intro}
\end{figure}

Researches on battery management systems (BMS) have received increasing attention for the rapid commercialization of electric vehicles (EVs) \cite{wang_energy_2019}. One of the core tasks of BMS is to predict State-of-Health (SoH) of Li-ion batteries. SoH is defined as the ratio of the current releasable battery charge to its rated capacity. It gradually decreases after charging and discharging for a number of cycles, indicating a shrinkage in capacity and maximum power. However, this vital indicator cannot be measured directly due to the complex dynamic behavior and time-varying conditions of Li-ion batteries\cite{zou_novel_2023}. The task of predicting SoH is to estimate SoH of future charging-discharging cycles given aging data (current, voltage, temperature, etc.) within every historical cycle (see Fig. \ref{fig:intro}).

Accuracy guarantees safety, and safety protects life. BMS needs accurate SoH predictions to optimize energy consumption, prevent over-charging and over-discharging, and extend battery life. In contrast, inaccurate estimations of SoH may lead EVs to spontaneous combustion or anchoring.

Challenges in accurate SoH prediction can be concluded as the following three points: First, SoH is a highly-complicated non-linear function of current, voltage, temperature and other parameters of historical cycles. Theoretically, deep neural network is a perfect choice to fit this function and learn the aging trend from historical data. But practically, it is hard for existing time sequence model to both learn long-term patterns among different cycles and extract battery features within each cycle. Second, different from many application scenarios of artificial intelligence, the aging data of Li-ion batteries is scarce. This usually causes under-fitting on light weight models and over-fitting on larger models. Third, the charging and discharging behavior of one battery may significantly differ from another, even if they are of the same type. A domain gap exists between batteries working in different conditions.
Previous works on SoH estimation generally utilize models based on RNN or LSTM\cite{9137406, fan_novel_2020}. Though more efficient, these models suffer severely from forgetting long-term patterns \cite{shen_state_2022}. More importantly, they quickly forget patterns learnt from training set when being fine-tuned on a test battery, and thus perform poorly when data for fine-tuning is scarce. Recently, several transformer-based methods have been proposed\cite{gu_novel_2023, fu_transfer_2022, biggio_dynaformer_2022}. In order to convert initial data into a transformer style input, most of them use a CNN-based feature extractor to extract intra-cycle features. However, the input data within each cycle does not have a hierarchical waveform structure. Therefore, it is hard for convolution filters to capture intra-cycle features effectively. Additionally, this pipeline compresses all sample points within a cycle into one single dimension, which might induce feature loss.

In this work, we present CyFormer, a novel generalized cyclic time sequence model, to address the aforementioned problems. This CNN-free model follows the typical encoder-decoder pipeline of transformer. The encoder first extracts cyclic features from historical data and transmits them into the decoder. Then the SoH queries cross-attend to these features in the decoder and forms the final predictions. At the core of our model lies the cyclic attention (i.e., row-wise and column-wise attention) blocks. Row-wise attention block aims at extracting intra-cycle connections, whereas column-wise attention block aims at extracting inter-cycle connections (Fig. \ref{fig:intro}). Compared with CNN, these two blocks enable the encoder to capture connections between sample points in different cycles, and preserve intra-cycle features at the same time.


Extensive experiments demonstrate both the effectiveness and the robustness of CyFormer on SoH prediction. Compared with previous works, we adopt a more challenging criterion for testing to fully demonstrate the transfer learning performance of our model. More specifically, we fine-tune the model with only 10\% of the SoH data at the beginning, predict SoH of the remaining 90\% hidden cycles, and calculate prediction error with the ground truth of these hidden cycles. With this testing method, our model achieved an MAE of 0.75\% and an MAPE of 0.90\%, surpassing baseline methods by a large margin. Industries may acquire the first few cycles of SoH data in quality control process of every new battery. Our result means that industries can use this SoH data to train a model which could give highly accurate predictions of SoH on BMS.

Our contributions can be concluded as follows:

\begin{itemize}
    \item We proposed CyFormer, a generalized cyclic time sequence model with row-wise and column-wise attention mechanism. With CyFormer, we gained highly accurate SoH predictions and achieved new SOTA in SoH estimation.
    \item We adopt a transfer learning style testing criterion, which is closer to the real application scenario. Experiments showed that our model maintains accuracy by this criterion.
    \item We designed a light weight version of CyFormer for BMS by pruning unnecessary modules. It becomes significantly more efficient than the initial version with only a tiny loss on accuracy.
\end{itemize}

\section{Related Work}
\paragraph{SoH Prediction}
Modern SoH prediction methods can be generally divided into three categories: direct measurement methods, adaptive algorithms and data-driven methods\cite{gu_novel_2023}. Direct measurement methods \cite{hu_battery_2016, zheng_novel_2019} analyze the aging behavior through numerous laboratory tests. These off-line methods require specialized sensors in laboratories. Adaptive algorithms use traditional mathematical models and numerical filters. Lim et al. \cite{lim_fading_2016} proposed Fading Kalman filter (FKF), which avoids large estimation errors in conventional Kalman filter. This method induces large computational costs \cite{shen_state_2022}, and is not efficient enough to be deployed on BMS.

Data-driven methods can be further divided into machine learning methods and deep learning methods. Machine learning methods typically utilize support vector machine (SVM) \cite{meng_lithium-ion_2018, tan_online_2021, cai_multiobjective_2020} or Gaussian process regression (GPR) \cite{yang_novel_2018, liu_remaining_2019, liu_data-driven_2021}. Most existing SOTAs adopt deep learning methods, such as CNN-LSTM \cite{9137406}, ViT \cite{fu_transfer_2022} or DynaFormer \cite{biggio_dynaformer_2022}. Fan et al. \cite{fan_novel_2020} proposed a hybrid neural network, which extracts local information with CNN and captures time dependencies with GRU. To better capture global representation, Gu et al. \cite{gu_novel_2023} proposed a CNN-Transformer framework that replaces recurrent modules with transformer encoder and decoder. To reduce oscillations, Shen et al. \cite{shen_state_2022} introduced Immersion and Invariance (I\&I) adaptive observer into the transformer-based pipeline.

\paragraph{Time Series Analysis}
In time series forecasting, one of the most prominent models is ARIMA \cite{box_recent_1968}. Flunkert et al. \cite{salinas_deepar_2020} first integrated auto-regression with RNN and proposed DeepAR, a probabilistic forecasting network. Bai et al. \cite{bai_empirical_2018} discovered that a simple convolutional architecture outperforms canonical recurrent networks (e.g., RNN, LSTM) on a wide spectrum of tasks and datasets. Li et al. \cite{li_enhancing_2019} proposed LogSparse Transformer with only $O(L(log L)^2)$ memory cost. They also utilized convolutional self-attention so that local context can be better incorporated into attention mechanism.


\paragraph{Transfer Learning}
In many deep learning tasks, a domain gap exists between training and testing datasets. Therefore, many generalized transfer learning methods have been proposed. Kumar et al. \cite{kumar_fine-tuning_2022} suggested LP-FT, a two-step strategy that first trains linear probing module and then fine-tunes the entire model. Similar strategies have been adopted in SoH predictions. To boost performance on batteries in different working conditions, Fu et al. \cite{fu_transfer_2022} conducted fine-tuning with SoH data of the first few cycles to shift the model to the testing battery.


\section{Task Statement}\label{sec:task}
As illustrated in Fig. \ref{fig:intro}, consider that we have a battery working at the end of the t-th charge-discharge cycle. Given size of the prediction window $n_{out}$, the task of our model is to predict SoH value of cycle $t+1$ to $t+n_{out}$, based on the aging data of cycle $1$ to $t$. To be specific, assume that $l_{sample}$ is the sample point number within each cycle, and $c$ is the number of physical quantities measured at each sample point (i.e., input channel size). The input of this task can be organized as the following $t \times l_{sample}$ matrix, 
\begin{align}\label{fml:input}
input=\left[\begin{array}{cccc}
X_{11}, & X_{12},& \cdots, & X_{1l_{sample} } \\
X_{2
1}, & X_{22},&  \cdots, & X_{2l_{sample} } \\
\vdots & \vdots &  & \vdots\\
X_{t1}, & X_{t2},&  \cdots, & X_{tl_{sample} }
\end{array}\right]
\end{align}
where each $X_{ij}$ is a vector of size $C$. It is composed of physical quantities like current, voltage and temperature. The output of this task is a sequence of predicted SoH values, 
\begin{align}
output=\left\{SoH_{i}|i=t+1,\cdots, t+n_{out}\right\}
\end{align}
where $SoH_{i}$ is a number within [0, 1].


There are two special cases worth considering. The first one is named just-in-time (JIT) prediction, where we set $n_{out}$ as 1 and only predict SoH of the current cycle. The second one is named Remaining Useful Life (RUL) prediction, where we set $n_{out}$ to a pre-defined maximum and predict the cycle number when SoH decreases to a certain threshold. This threshold represents the scrapping point of batteries, and the number of remaining cycles indicates the remaining life of the battery. In this work, we focus on JIT research.

\section{Method}

\begin{figure*}[htbp]
\centerline{\includegraphics[width = \textwidth]{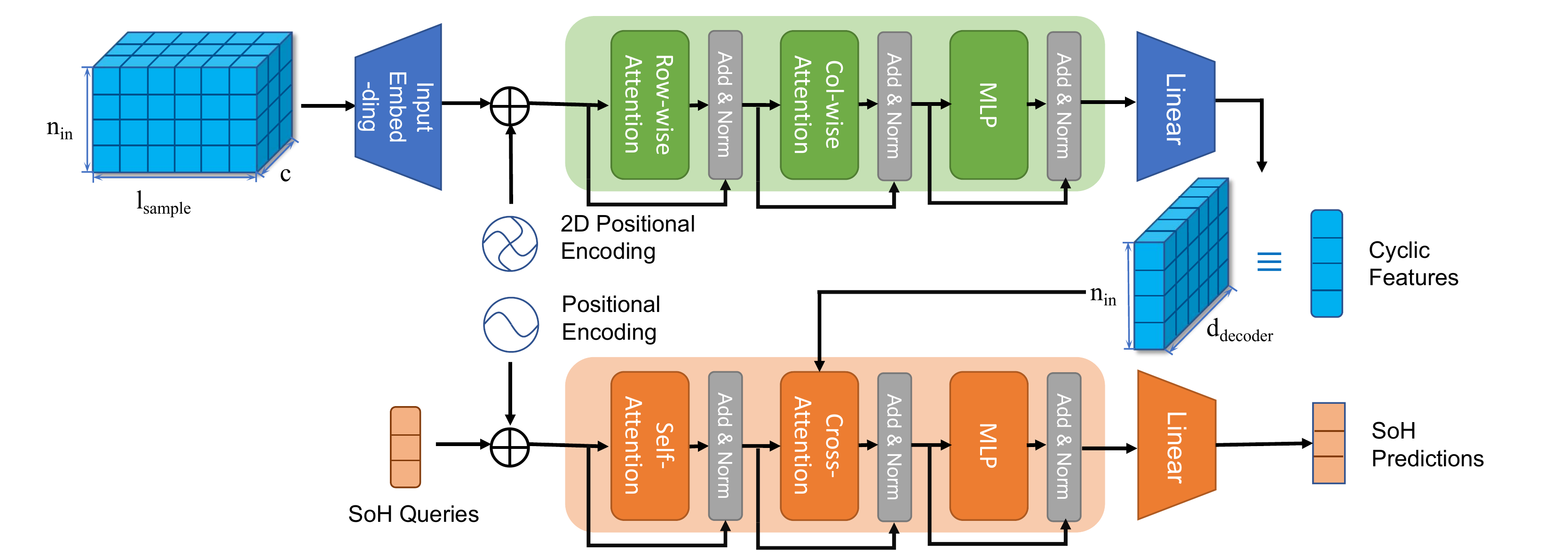}}
\caption{An overview of our model. Components of the encoder are colored green, whereas componets of the decodered are colored orange. Each cube in the input or square in the output represents a number, while each square in rounded rectangles represents a vector of size $d_{decoder}$.}
\label{fig:1}
\end{figure*}

An overview of our model is depicted in Fig.\ref{fig:1}. As described in Section \ref{sec:task}, our model directly processes two-dimensional cyclic data, rather than the one-dimensional input sequence of the classical transformer architecture. The encoder first encodes the two-dimensional input into a one-dimensional feature sequence, and then feed it into the decoder. In the decoder, the randomly-initialized SoH queries cross-attend to these features to form prediction values of $n_{out}$ future cycles. In the following sections, we will discuss the detailed structure of the encoder and the decoder respectively.

\begin{figure}[!t]
\removelatexerror
    \begin{algorithm}[H]
    \caption{Row-wise and Column-wise Attention}
        \begin{algorithmic}[0]
        \label{alg:attn}
            \REQUIRE Matrix $[X_{ij}^{'}]$ with channel size c 
    	\ENSURE Matrix $[O_{ij}]$ with channel size c 
            \FOR {each row $R_{i}$ in $[X_{ij}^{'}]$}
                \STATE {// Input projections}
                \STATE $Q_i, K_i, V_i \leftarrow linear(R_{i})$
                \STATE {// Self-attention}
                \STATE $A_i \leftarrow softmax\left(\frac{1}{\sqrt{d_{encoder}}} \mathbf{Q}_{i}^{\top} \mathbf{K}_{i}\right)$
                \STATE $R_i\leftarrow layerNorm(R_{i} + V_iA_i^{\top})$
            \ENDFOR
    	\FOR {each column $C_{j}$ in $[X_{ij}^{'}]$}
                \STATE {// Input projections}
                \STATE $Q_j, K_j, V_j \leftarrow linear(C_{j})$
                \STATE {// Self-attention}
                \STATE $A_j \leftarrow softmax\left(\frac{1}{\sqrt{d_{encoder}}} \mathbf{Q}_{j}^{\top} \mathbf{K}_{j}\right)$
                \STATE $C_j\leftarrow layerNorm(C_{j} + V_jA_j^{\top})$
            \ENDFOR
            \STATE $[O_{ij}]\leftarrow[X_{ij}^{'}]$
        \end{algorithmic}
    \end{algorithm}
\end{figure}

\subsection{Encoder}

The encoder is composed of four main parts, namely the input embedding module, the 2D positional encoding module, a stack of encoder layers and the output head. 

Both the input embedding and the output head are linear layers. The input embedding module extends the channel size of each input token to $d_{encoder}$. It applies the following affine transformation on each vector in the input matrix \ref{fml:input}:
\begin{align}
X_{ij}^{'} = W^{\top}X_{ij} + b
\end{align}
where $W^{T}$ is a $C \times d_{encoder}$ weight matrix and b is a bias vector of size $d_{encoder}$. As the output of the embedding module, $X_{ij}^{'}$ is then fed into the first encoder layer.

The output head fully connects all sample points in each input cycle to form $n_{in}$ feature vectors with the channel size of $d_{decoder}$. It can be described as Formula \ref{fml:enout}.
\begin{align}\label{fml:enout}
F_{i}=\sum_{m=1}^{l_{sample}} O_{m}^{\top} X_{i m}+b
\end{align}
where $F_{i}$ is the i-th vector in the cyclic feature sequence. Each $W_{m}^{T}$ is a $d_{encoder} \times d_{decoder}$ weight matrix, and b is a bias vector of size $d_{decoder}$.


The 2D positional encoding of each input token is defined as: 

\begin{align}
PE2D_{(cycle, sample)}=PE1D_x + PE1D_y
\end{align}

Where $PE1D_x$ and $PE1D_y$ are both 1D sinusoidal positional encodings calculated by \cite{vaswani2017attention}:

\begin{align}
PE1D_{(pos,2i)} = sin(pos/10000^{2i/d_{model}}) \label{fml:1}\\
PE1D_{(pos,2i+1)} = cos(pos/10000^{2i/d_{model}}) \label{fml:2}
\end{align}

Each encoder layer consists of a row-wise attention block, a column-wise attention block and a 3-layer MLP. Each of these blocks is followed by a residual block and a layer normalization block. Both row-wise and column-wise attention blocks derive from self-attention blocks (Alg. \ref{alg:attn}). They are designed to capture intra-cycle and inter-cycle connections respectively.

\begin{figure}[htbp]
\centerline{\includegraphics[width = 0.5\textwidth]{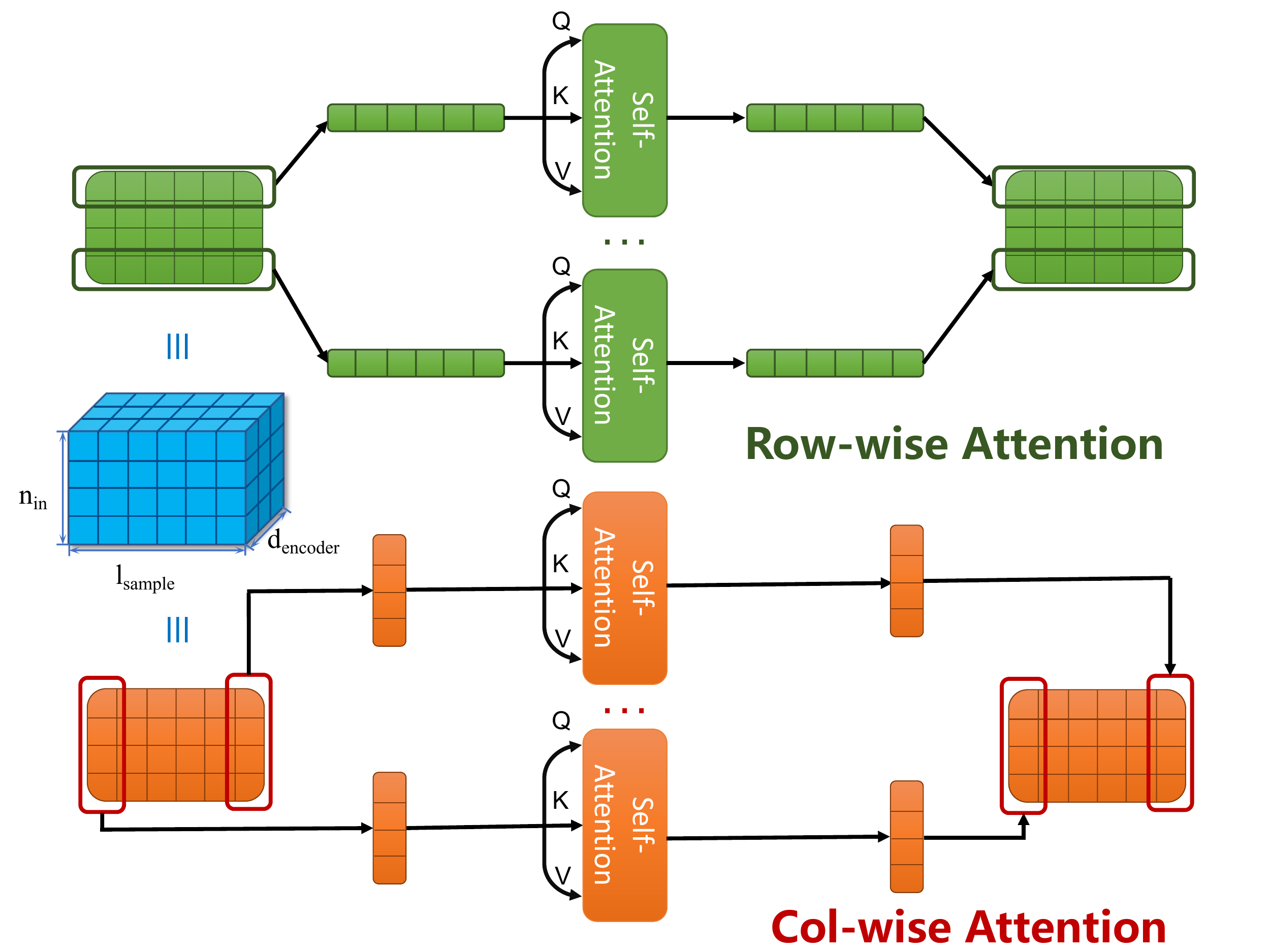}}
\caption{An illustration of cyclic attention mechanism. The upper part and the lower part show the structure of row-wise and column-wise attention blocks, respectively. Each square in rounded rectangles represents a vector of size $d_{encoder}$.}
\label{fig:2}
\end{figure}

\paragraph{Row-wise Attention}
The upper part of Fig.\ref{fig:2} illustrates the structure of a row-wise attention block. Row-wise attention aims at capturing connections between data sampled at different times within a single cycle. To this end, we first split the two-dimensional input into single rows. Each row contains all sample points within a particular cycle. Then each row is regarded as an individual input sequence and goes through a self-attention block with shared weights. To be specific, each row generates its own queries, keys and values. Following the typical multi-head attention mechanism, we first calculate dot-product affinities between queries and keys to form attention weights. Then we multiply the values with their corresponding weights, and feed the result into an output linear layer. After going through self-attention blocks, the outputs of all rows are concatenated to form a final output with the same shape as the original input.

\paragraph{Column-wise Attention}
The lower part of Fig.\ref{fig:2} illustrates the structure of a column-wise attention block. Column-wise attention aims at capturing connections between data sampled at the same time but in different cycles. Similar to row-wise attention, we achieved this goal by slicing the input data into individual columns, feeding them into self-attention blocks with shared weights, and concatenating them into an output matrix.

It should be noticed that the inputs and outputs of row-wise and column-wise blocks all have the same shape, which renders these blocks compatible with the sequential transformer architecture. Additionally, our method preserves intra-cycle features since the second dimension is never squeezed until the end of the encoder.

\subsection{Decoder}

The decoder is composed of three main parts, namely the positional encoding module, a stack of decoder layers, and a linear output head. A sequence of randomly-initialized SoH queries goes through these parts to form the final SoH predictions. The query sequence consists of $n_{out}$ vectors with a channel size of $d_{decoder}$:
\begin{align}
query=\left[\begin{array}{cccc}
Q_{1}, & Q_{2},& \cdots, & Q_{n_{out}}
\end{array}\right]
\end{align}
The output head is a linear layer. It fully connects all channels of each query vector and outputs $n_{out}$ SoH values:
\begin{align}
SoH_{t+i} = W^{\top}Q_{i} + b
\end{align}
In the decoder, we use the 1D sinusoidal positional encoding defined in Formula \ref{fml:1} and \ref{fml:2}.

Following the typical transformer decoder architecture, each decoder layer consists of a self-attention block, a cross-attention block and a 3-layer MLP. Similar to the encoder, each of these blocks is followed by a residual block and a layer normalization block. The cross-attention block informs queries of the historical features $\left\{F_{i}\right\}$, while the self-attention block keeps the historical trend among all the predictions.


\section{Experiment}
In this section, we first introduce the dataset and the data pre-processing procedure (Sec. \ref{sec:dataset}). Then, We introduce evaluation metrics (Sec. \ref{sec:criteria}) and implementation details (Sec. \ref{sec:detail}). We conducted a comparative experiment with CNN-LSTM and CNN-Transformer (Sec. \ref{sec:compare}) to demonstrate the accuracy of CyFormer. To better validate each component of our model, we provide detailed ablation studies (Sec. \ref{sec:abalation}). Finally, we present a light weight version of CyFormer, striking a balance between accuracy and computational costs(Sec. \ref{sec:pruning}).

\subsection{Dataset}\label{sec:dataset}
\begin{figure}[htbp]
    \centerline{\includegraphics[scale=0.35]{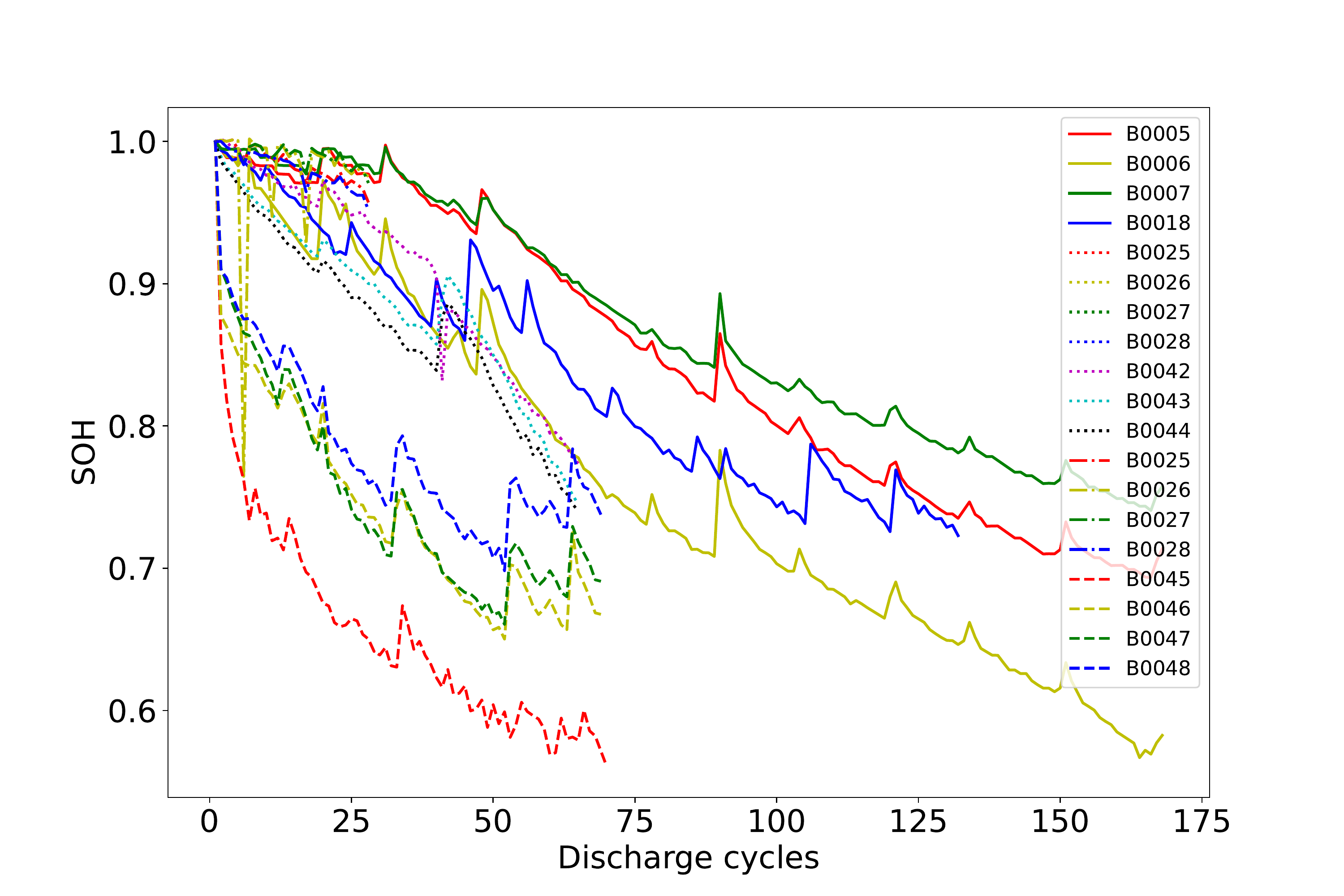}}
    \caption{The SoH decay curves of the 19 batteries in the Battery Data Set. The same linestyle indicates the same group of batteries. Batteries are marked with different colors in the same group. }
    \label{fig:Data}
\end{figure}

 We carry out the experiment with the Battery Data Set provided by NASA Ames Prognostics Center of Excellence \cite{dataset, data}. We removed batteries that are extremely inconsistent with the common aging patterns of batteries. Fig. \ref{fig:Data} shows the aging curve of the nineteen batteries we used. These battery cells worked in different ambient temperatures($4 \tccentigrade$, $24 \tccentigrade$, $43 \tccentigrade$).In each cycle, they were first charged through a constant current - constant voltage (CC-CV) procedure with the upper voltage at 4.2V until the current decayed to 20 mA. Then they were discharged with constant or pulse current waveforms until each of the cells reached its cut-off voltage. We use load voltage (Vl), load current (Cl), measured voltage (Vm), measured current (Cm) and temperature (T) curves of the discharge process (see Fig. \ref{fig:intro}). The SoH of a batteryis defined as the ratio of the maximum charge to its rated capacity:
\begin{align}
\mathrm{SOH}=\frac{Q_{\max }}{C_r} \times 100 \% 
\end{align}
where $Q_{\max}$ is the maximum charge available from the current battery and $C_r$ is the rated capacity. To align sample points in different cycles, we linearly interpolated and re-sampled intra-cycle data.

Among all nineteen batteries, one is selected as the target battery for testing, and others are used as the source dataset for training. To this end, aging data of the target battery is further divided into the fine-tuning segment(10\%) and the hidden segment(90\%). We adopt a two-stage transfer learning strategy to narrow the domain gap between the source and the target batteries. We first train the model on the source dataset to fit general working conditions, and then fine-tune the model on the fine-tuning segment of the target battery to shift to a new working condition. Finally, evaluation of the model is conducted on the hidden segment of the target battery.


\subsection{Evaluation Metrics}\label{sec:criteria}
To evaluate the performance of CyFormer, three different evaluation metrics are employed: mean absolute percentage error (MAPE), mean absolute error (MAE) and root mean square error (RMSE). MAPE represents the relative percentage error between the prediction and the actual value. MAE is the average of the absolute difference between the estimation and the actual value of SOH. It aims at measuring the average magnitude of errors of the proposed method. RMSE indicates the deviation of the estimation value and the actual value, and thus represents the quality of estimation. MAE, MAPE, RMSE are defined as:
\begin{align}
\text { MAE } & = \frac{1}{n} \sum_{i=1}^{n}\left|\widehat{y}_{i}-y_{i}\right| \\
\text { MAPE } & = \frac{1}{n} \sum_{i=1}^{n}\left|\frac{\widehat{y}_{i}-y_{i}}{y_{i}}\right| \\
\text { RMSE } & = \sqrt{\frac{1}{n} \sum_{i=1}^{n}\left(\widehat{y}_{i}-y_{i}\right)^{2}} 
\end{align}
where $y_{1}, y_{2}, \cdots, y_{n}$ are the actual values, $\widehat{y}_{1}, \widehat{y}_{2}, \cdots, \widehat{y}_{n} $ are the predicted values, and $n$ is the number of testing samples.

\subsection{Implementation Details}\label{sec:detail}
We choose MAE as the loss function. The network is trained with Adam optimizer with a learning rate of 0.0001. We set $\beta_1$ as 0.9 and $\beta_2$ as 0.999. Grid search is used to obtain the optimal model parameters. The selected parameters are shown in Table \ref{table:Hyperparameter}.


\begin{table}[htbp]
\begin{center}
	\caption{Hyperparameter settings}
	\label{table:Hyperparameter}
	\begin{tabular}{ll}
		\hline\hline\noalign{\smallskip}	
		Hyperparameters & value\\
		\noalign{\smallskip}\hline\noalign{\smallskip}
        $l_{sample}$ & 32 \\
        $d_{encoder}$ & 16\\
        $d_{decoder}$ & 16\\
        $n_{in}$ & 16 \\
        Gamma   &  0.1 \\
        Batch size & 32 \\
        Epochs & 1500 \\
        Encoder layers & 4   \\
        Decoder layers & 4  \\
        Attention heads  $h$  & 8 \\
        Learning rate(training) &  0.0001  \\
        Learning rate(fine-tuning) &  0.0002  \\
		\noalign{\smallskip}\hline
	\end{tabular}
\end{center}
\end{table}

The input window size $n_{in}$ is defined as the number of cycles contained in the input sequence. The SoH prediction performances with different input window size $n_{in}$ are shown in \ref{table:in}. When $n_{in}$ increases, the accuracy increases simultaneously, whereas the computational cost rises as well. Additionally, ground-truth data for fine-tuning would be scarce if $n_{in}$ were too large. Therefore, we set $n_{in}=16$ as it strikes a balance between accuracy and efficiency.
\begin{table}[htbp]
\begin{center}
	\caption{The SoH result of different $n_{in}$}
	\label{table:in}
	\begin{tabular}{cccccc}
		\hline\hline\noalign{\smallskip}	
		$n_{in}$ &  FLOPs & Params & MAE & MAPE & RMSE \\
		\noalign{\smallskip}\hline\noalign{\smallskip}
		8          & 0.09 & 0.32 & 2.68\% & 2.95\% & 3.13\%  \\
        12         & 0.13 & 0.33  & 1.56\% & 1.66\% &1.90\%  \\
        16         & 0.17 & 0.35 & 0.75\% & 0.89\% & 0.95\%  \\
        32         & 0.34 & 0.41 & 0.73\% & 0.90\% & 0.95\%  \\
		\noalign{\smallskip}\hline
	\end{tabular}
\end{center}
\end{table}


We adopt a transfer learning style testing criterion. The fine-tuning segment of the target battery only contains a small amount (10\%) of data at the beginning of the ageing phase. The feature extraction modules are mainly trained on the source dataset. In the fine-tuning process, parameters in the decoder are mainly modified, and the model quickly shifts to the target domain with the 10\% fine-tuning segment.

\begin{figure}
        \centering
        \scriptsize
        \begin{tabular}{ccc}
                \includegraphics[width=9cm]{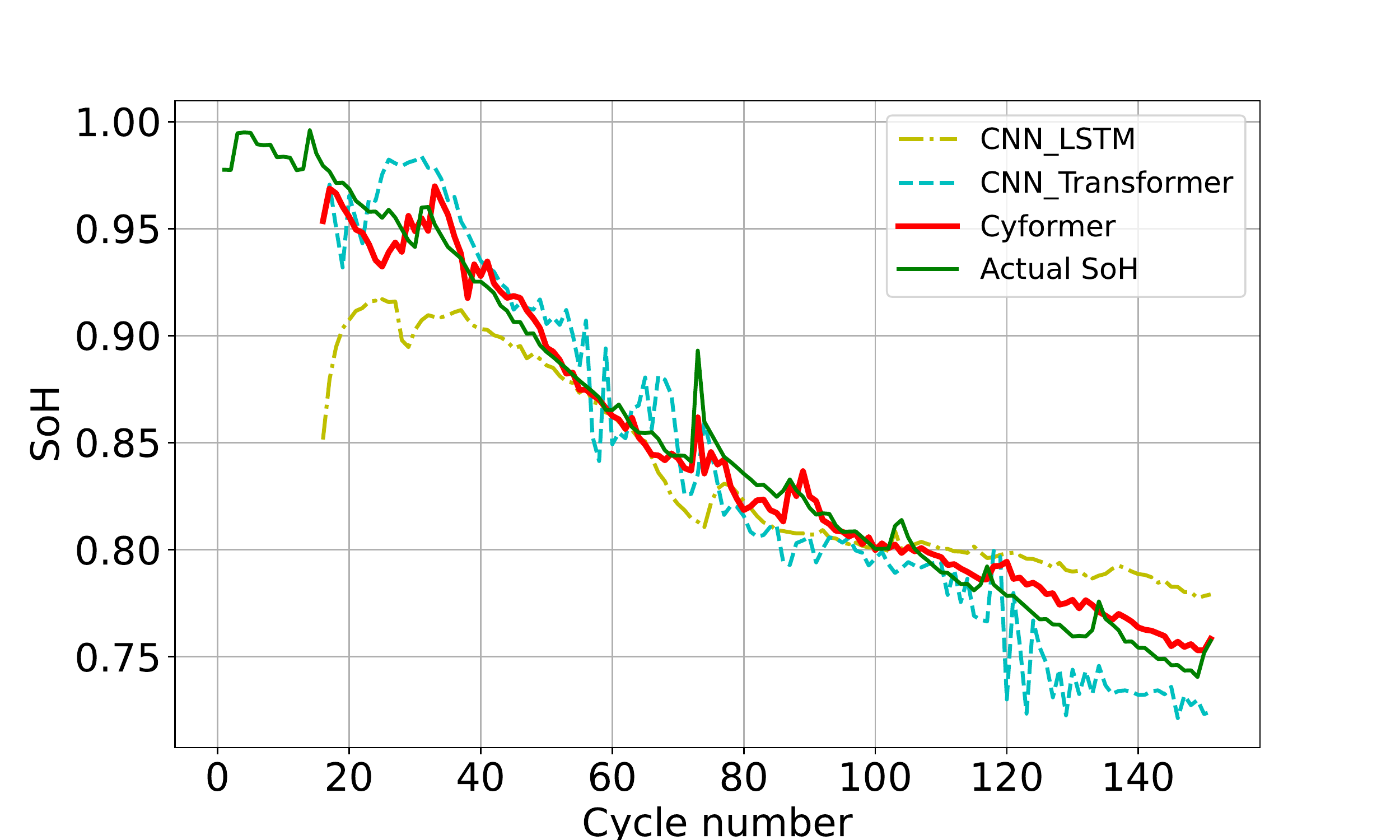} \\    
                (a) \\
                \includegraphics[width=9cm]{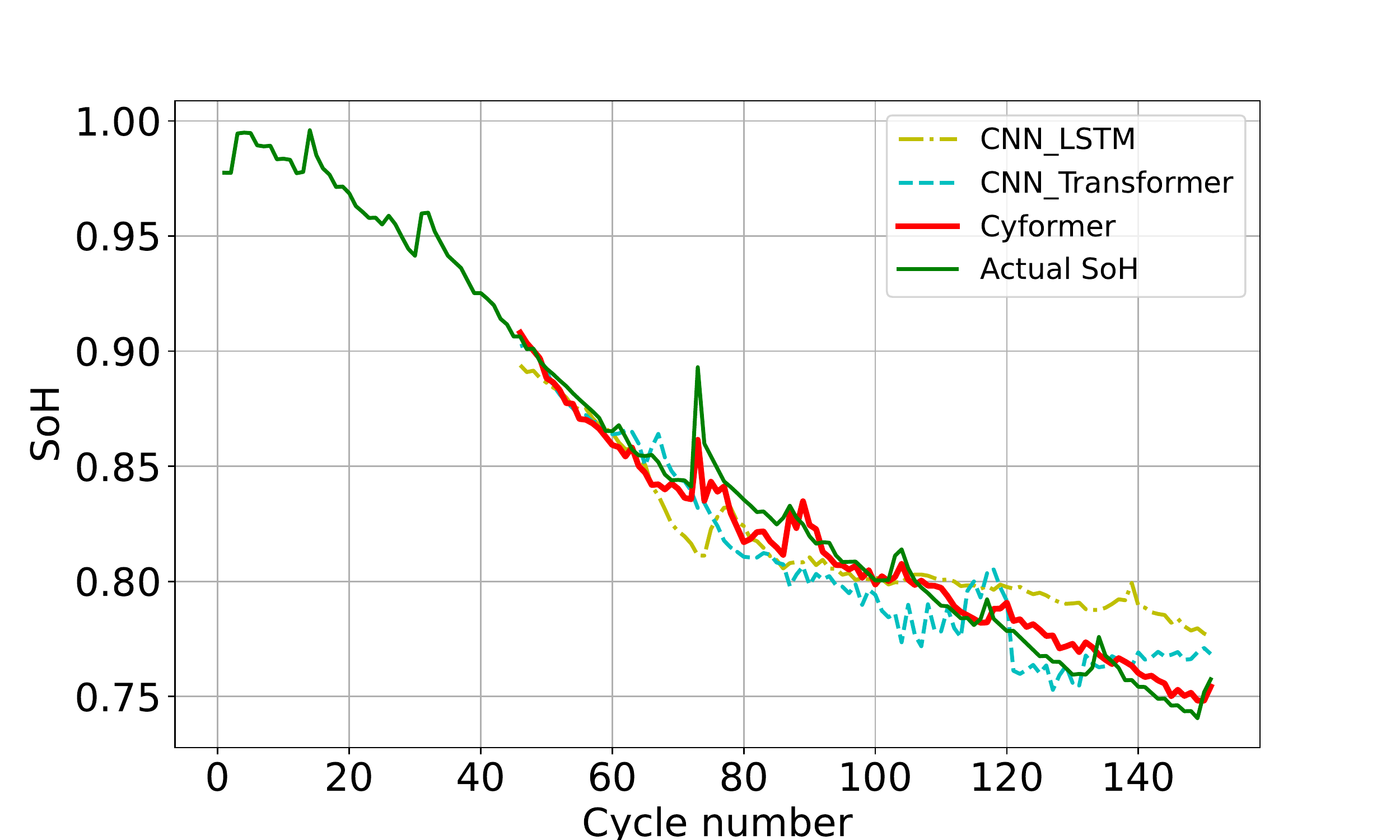} \\
                (b)\\
                \includegraphics[width=9cm]{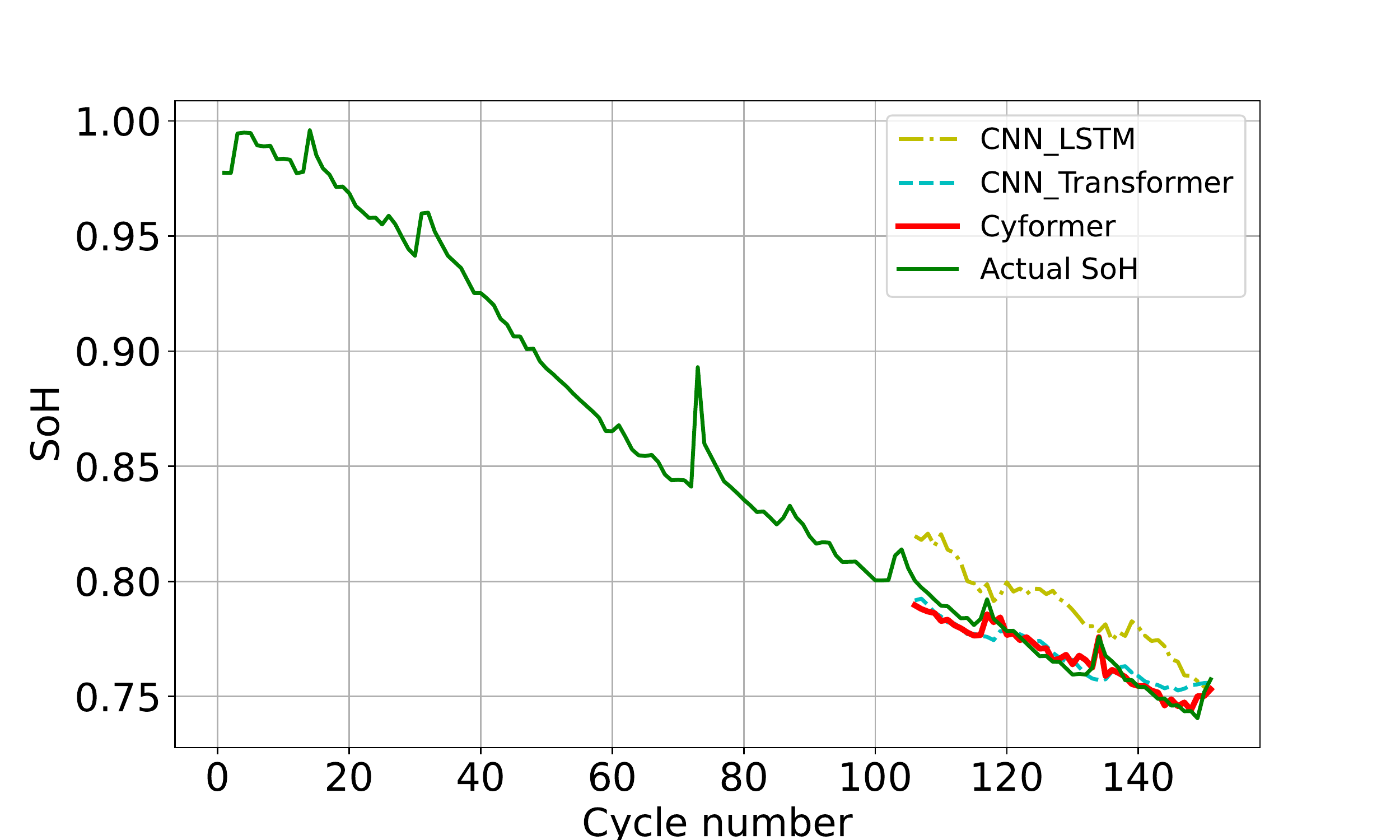} \\
                (c)\\
        \end{tabular}
        \caption{The result of SoH estimation based on 10\% (a), 30\% (b), 70\% (c) transfer learning dataset.}
        \label{fig:test}
\end{figure}

\subsection{Comparison with Other Methods}\label{sec:compare}
As mentioned in section \ref{sec:dataset}, Cell \#B0007 is randomly selected as the target battery. Other batteries are selected as the source dataset. The first 10\%, 30\%, or 70\% of \#B0007 were adopted for offline fine-tuning, while the remaining part (90\%, 70\%, or 30\%) were used for online evaluation. As typical baselines, CNN-LSTM and CNN-Transformer were employed to estimate battery SoH with the same testing criteria.

As shown in Tab. \ref{table:compare}, the MAEs, MAPEs and RMSEs of the CyFormer are within 1\%, while for CNN-LSTM and CNN-Transformer, the highest errors are about 4\% and 3\%, respectively. CyFormer achieves the lowest loss among all three methods under all three circumstances, demonstrating its effectiveness and robustness. It should be noticed that CyFormer can achieve accurate prediction using only 10\% or less fine-tuning data, while CNN-Transformer need at least 70\% to reach a comparable result. This can also prove 
the transfer learning efficiency of CyFormer.




\begin{table}[htbp]
\begin{center}
	\caption{Comparison of estimation errors among CyFormer and other methods}
	\label{table:compare}
	\begin{tabular}{lllll}
		\hline\hline\noalign{\smallskip}	
		Battery & Method & MAE & MAPE & RMSE \\
		\noalign{\smallskip}\hline\noalign{\smallskip}
		B0007(10\%) 
		  &  CNN-LSTM & 2.69\% & 2.98\% & 3.30\% \\
            &  CNN-Transformer  & 2.01\%  & 2.41\% & 2.33\% \\
            & CyFormer & \textbf{0.75\%} & \textbf{0.89\%} &\textbf{0.95\%}  \\
        B0007(30\%) 
		  &  CNN-LSTM & 1.74\% & 2.19\% & 2.17\% \\
            &  CNN-Transformer  & 1.12\%  & 1.56\% & 1.67\% \\
            & CyFormer & \textbf{0.66\%} & \textbf{0.87\%} & \textbf{0.96\%} \\
        B0007(70\%) 
		  &  CNN-LSTM & 1.70\% & 2.11\% & 1.89\% \\
            &  CNN-Transformer  & 0.66\%  & 0.86\% & 0.82\% \\
            & CyFormer & \textbf{0.38\%} & \textbf{0.52\%} & \textbf{0.49\%}  \\
		\noalign{\smallskip}\hline
	\end{tabular}
\end{center}
\end{table}

Fig. \ref{fig:test} (a)-(c) show the SoH prediction results on the target battery (\#B0007). The SoH prediction accuracy of CyFormer surpasses other methods by a large margin, especially when the fine-tuning proportion is 10\% (Fig. \ref{fig:test}(a)). When the fine-tuning proportion expands to 30\%, CNN-LSTM and CNN-Transformer models closely follow the battery ageing trend only in the first twenty cycles. Significant improvement on accuracy of CNN-Transformer has not appeared until the fine-tuning proportion reaches 70\%. In contrast, the CNN-LSTM model still jitters significantly even if the the fine-tuning proportion reaches 70\%.

\subsection{Ablation Study}\label{sec:abalation}
In order to validate the effect of row-wise and column-wise attention blocks, we conducted abalation studies on following conditions.
\begin{itemize}
    \item w/o Row-wise: CyFormer without row-wise structure
    \item w/o Col-wise: CyFormer without column-wise structure
    \item w/o Row-wise + Col-wise: CyFormer without row-wise and column-wise structure
\end{itemize}

\begin{figure}[htbp]
    \centerline{\includegraphics[scale=0.38]{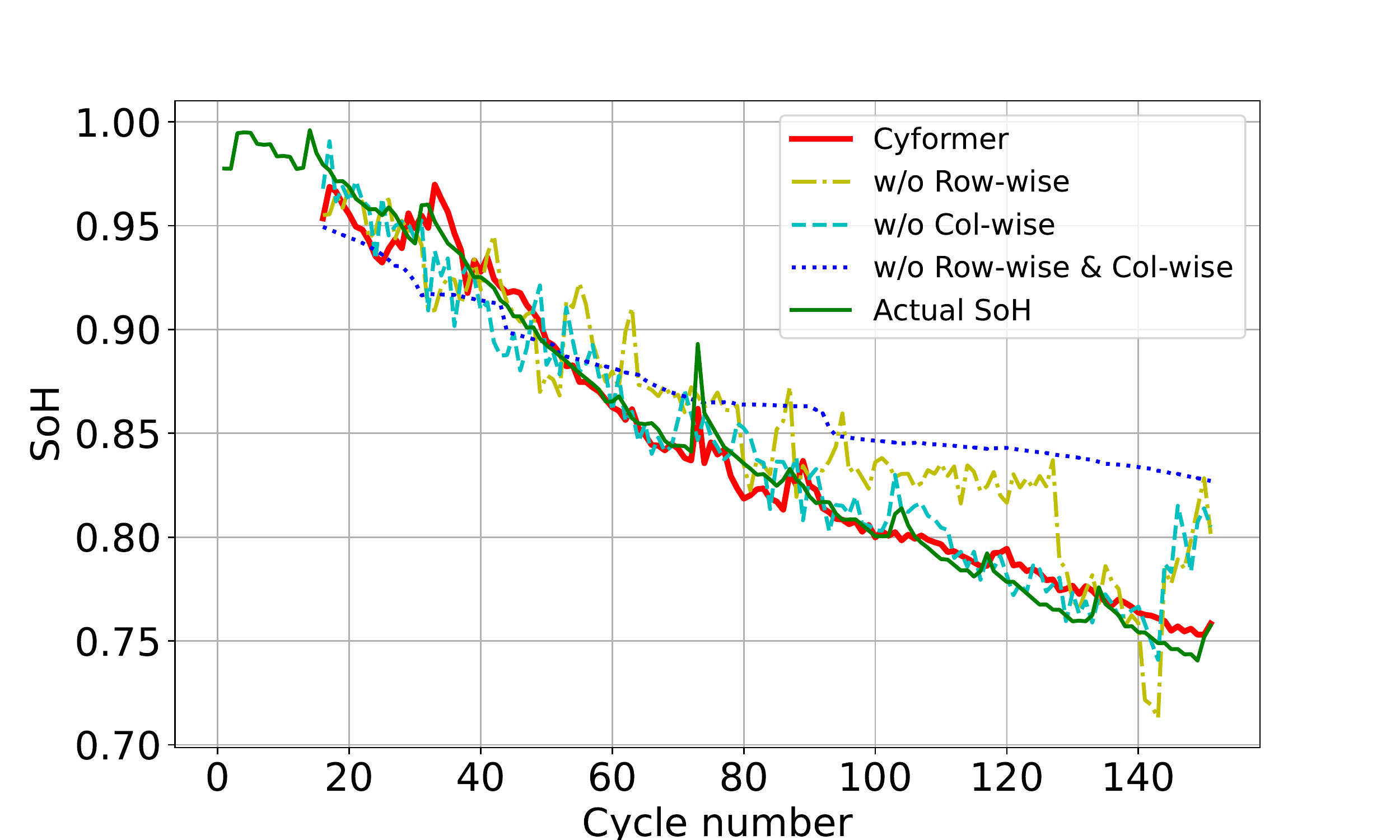}}
    \caption{Ablation Study}
    \label{fig:Ablation}
\end{figure}

According to Fig. \ref{fig:Ablation}, cutting off row-wise and column-wise attention blocks decreases accuracy, verifying the effectiveness of the two structures. As shown in Tab. \ref{table:Ablation}, the column-wise and row-wise atttention blocks reduce MAE by 3.06\%, MAPE by 3.82\%, and RMSE by 3.54\%.

\begin{table}[htbp]
\begin{center}
	\caption{Ablation study of column-wise and row-wise attention blocks}
	\label{table:Ablation}
	\begin{tabular}{cccccc}
		\hline\hline\noalign{\smallskip}	
		Col-wise. &  Row-wise. & MAE & MAPE & RMSE \\
		\noalign{\smallskip}\hline\noalign{\smallskip}
                 &          & {3.81\%} & {4.72\%} & {4.53\%} \\
		       & $\surd$ & {1.87\%} & {2.18\%}& {3.42\%}  \\
        $\surd$  &          & {2.93\%} & {3.58\%} & {4.02\%} \\
        $\surd$  & $\surd$  & \textbf{0.75\%} & \textbf{0.89\%} & \textbf{0.95\%} \\
		\noalign{\smallskip}\hline
	\end{tabular}
\end{center}
\end{table}

\subsection{Pruning}\label{sec:pruning}

In this section, we designed a light weight version of CyFormer by pruning. To be specific, we reduced the number of encoders (i.e., depth), and the number of sampling points. We only change one component each time to observe how that affects performance and efficiency.

\begin{figure}[htbp]
    \centerline{\includegraphics[scale=0.3]{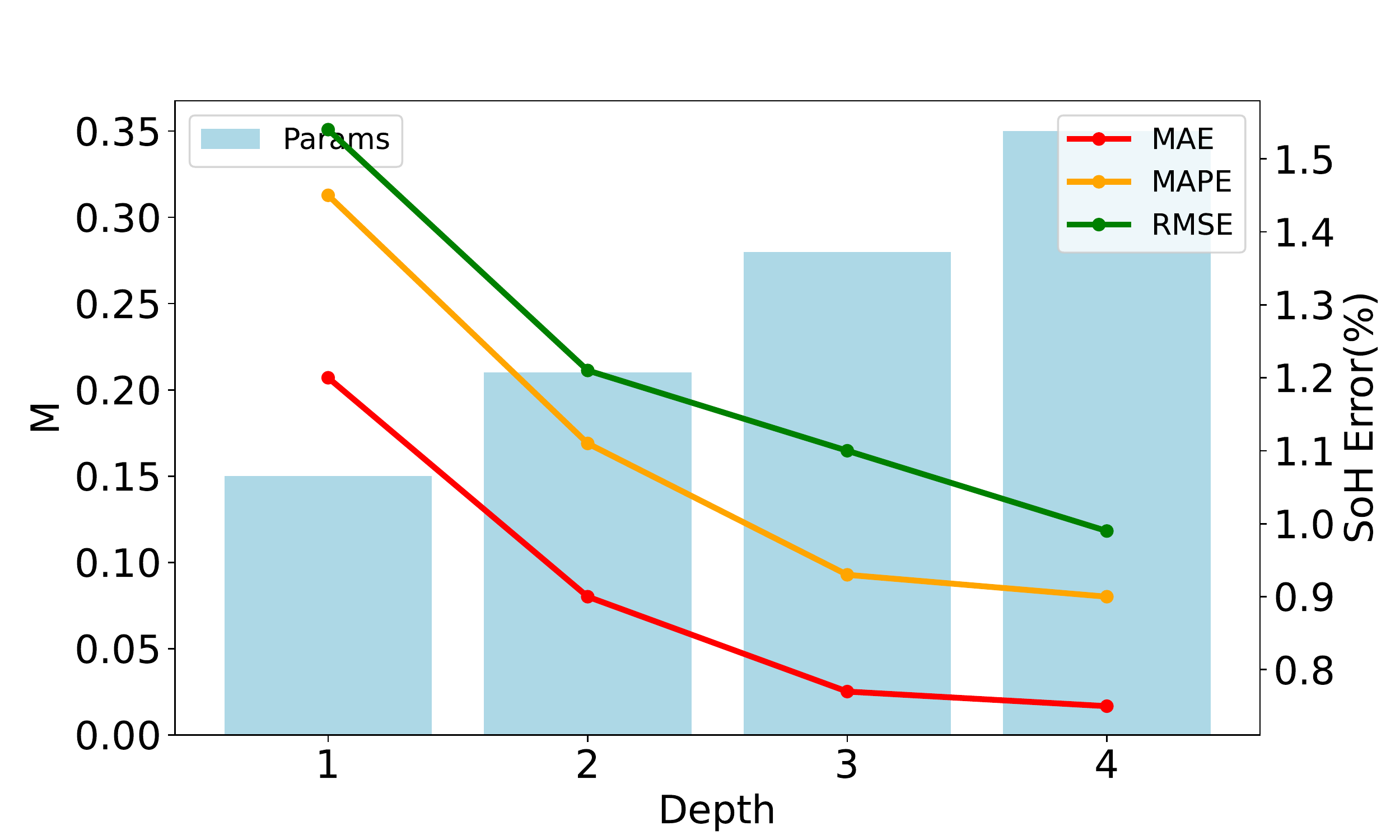}}
    \caption{The results of pruning study on model depth.}
    \label{fig:depth}
\end{figure}

\paragraph{Depth}
The depth of the model is defined as the number of CyFormer encoders. It is highly related to the effectiveness of feature extraction. The initial depth is set as 4. To make it more efficient, we carried out four groups of experiments that set depth to 1-4, respectively. The RMSEs, MAEs, MAPEs and FLOPs of SoH prediction are shown in Fig. \ref{fig:depth}.

It can be seen from Fig. \ref{fig:depth} that the feature extraction ability becomes stronger as the model depth increases. Each of the first three layers improves performance significantly, while additional layers brings minor improvement. In consideration of prediction accuracy and inference speed, we set the depth of the network as 3 in the light weight model.



\begin{figure}[htbp]
    \centerline{\includegraphics[scale=0.3]{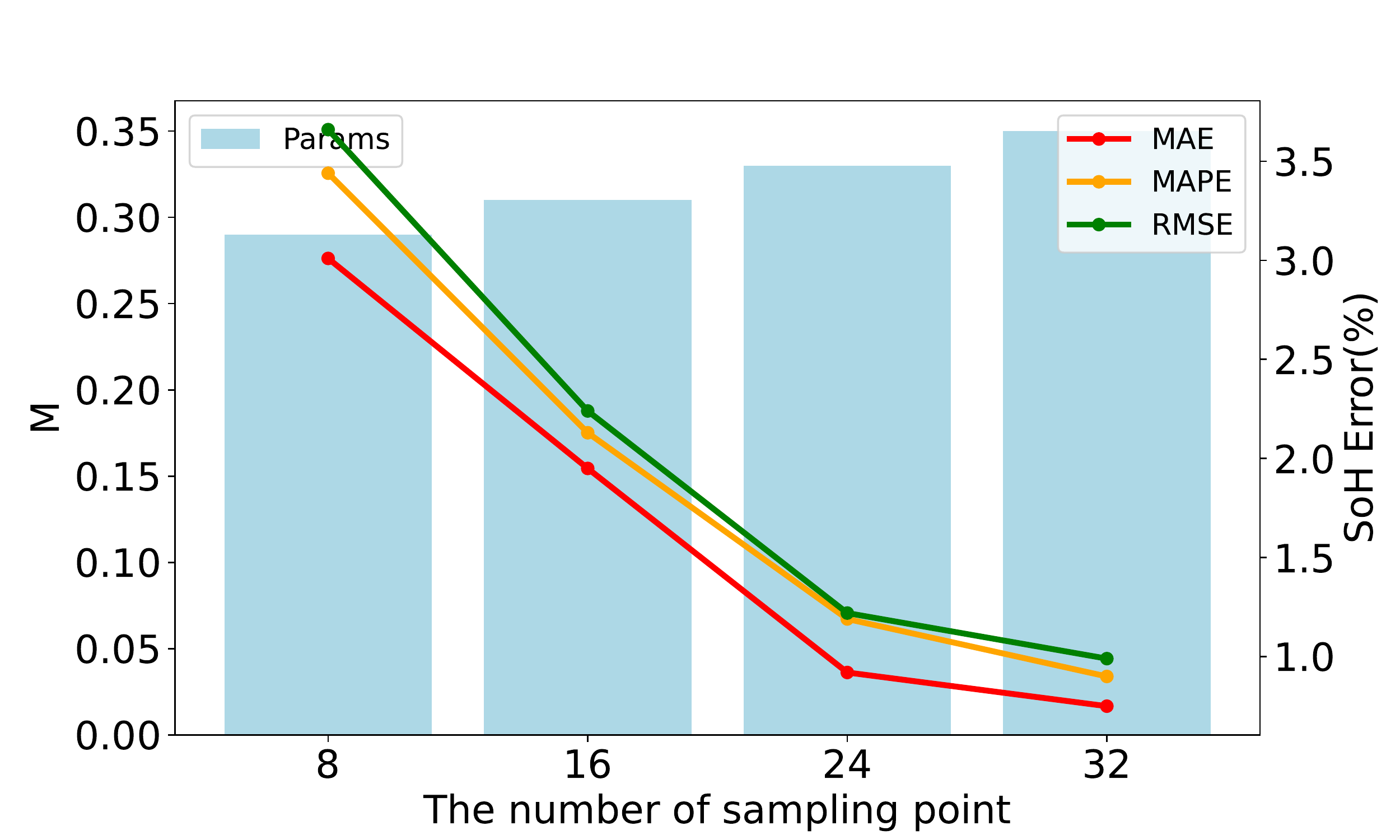}}
    \caption{The results of pruning study on model depth.}
    \label{fig:sample}
\end{figure}

\paragraph{Sampling Rate}
Fig. \ref{fig:sample} shows the loss and FLOPs when using different numbers of sampling points. We linearly interpolated and re-sampled from each cycle to form $l_{sample}$ sample points. Initially, we set $l_{sample}$ to 32. As $l_{sample}$ increases, the prediction accuracy augments, but FLOPs rises as well. Thus, we choose the elbow point 24 as $l_{sample}$, striking a balance between performance and computational costs.

The pruning results of each module are shown in Tab. \ref{table:pruning}. The joint effect of pruning both depth and sampling rate leads FLOPs and quantity of parameters to reduce by 41\% and 26\%, respectively. At the same time, the accuracy of the model is hardly affected.

\begin{table}[htbp]
\begin{center}
	\caption{Pruning Experiment}
	\label{table:pruning}
	\begin{tabular}{lccccc}
		\hline\hline\noalign{\smallskip}	
		Methods &  FLOPs & Params & MAE & MAPE & RMSE \\
		\noalign{\smallskip}\hline\noalign{\smallskip}
		Initial          & 0.17 & 0.35 & 0.75\% & 0.89\% & 0.95\%  \\
        Depth            & 0.13 & 0.28  & 0.77\% & 0.93\% &1.10\%  \\
        Sampling point   & 0.13 & 0.33 & 0.92\% & 1.19\% & 1.22\%  \\
        Overall          & 0.10 & 0.26 & 0.95\% & 1.17\% & 1.26\%  \\
		\noalign{\smallskip}\hline
	\end{tabular}
\end{center}
\end{table}

\section{Conclusion}
In this work, we present CyFormer, a generalized cyclic time sequence model with row-wise and column-wise attention mechanism. Via cyclic attention, our model effectively captures inter-cycle and intra-cycle connections. To narrow the domain gap among different working conditions, we adopt a two-stage transfer learning strategy. We also designed a light weight version of CyFormer for embedding systems by pruning. Experiments show that our model produces accurate SoH predictions using only 10\% data for fine-tuning, demonstrating the effectiveness and robustness of our model. CyFormer provides a potential solution for all cyclic time sequence prediction tasts, and we expect to see more applications of our method.


\section*{Acknowledgment}
This work was supported by General Terminal IC Interdisciplinary Science Center of Nankai University.

\vspace{12pt}
\end{document}